\documentclass[10pt,letterpaper]{article}

\usepackage{cogsci}

\cogscifinalcopy 

\usepackage{cmap}
\usepackage[T1]{fontenc}
\usepackage[american]{babel}
\usepackage{csquotes}
\usepackage{newtxtext,newtxmath}  
\usepackage{graphicx} 
\usepackage{float}  
\usepackage{xcolor}

\definecolor{TBblue}{HTML}{83A2D4}
\definecolor{FBorange}{HTML}{E0BAA1}
\definecolor{SAorange}{HTML}{E98C3C}

\usepackage[
  backend=biber,
  style=apa,
  natbib=true,
  annotation=false,
]{biblatex}
\usepackage{float} 

\usepackage[hidelinks]{hyperref}

\title{Simulated Annealing Enhances Theory-of-Mind Reasoning in Autoregressive Language Models}

\author[1,*]{\mbox{Xucong Hu (xuconghu@zju.edu.cn)}}
\author[2,*]{\mbox{Jian-Qiao Zhu (zhujq@hku.hk)}}
\affil[1]{Department of Psychology and Behavioral Sciences, Zhejiang University}
\affil[2]{Department of Psychology, The University of Hong Kong}
\affil[*]{Equal contribution}

\begin{document}

\maketitle

\begin{abstract}
Autoregressive language models are next-token predictors and have been criticized for only optimizing surface plausibility (i.e., local coherence) rather than maintaining correct latent-state representations (i.e., global coherence). Because Theory of Mind (ToM) tasks crucially depend on reasoning about latent mental states of oneself and others, such models are therefore often thought to fail at ToM. While post-training methods can improve ToM performance, we show that strong ToM capability can be recovered directly from the base model without any additional weight updates or verifications. Our approach builds on recent power-sampling methods \citep{karan2025reasoning} that use Markov chain Monte Carlo (MCMC) to sample from sharpened sequence-level (rather than token-level) probability distributions of autoregressive language models. We further find that incorporating annealing, where the tempered distribution is gradually shifted from high to low temperature, substantially improves ToM performance over fixed-temperature power sampling. Together, these results suggest that sampling-based optimization provides a powerful way to extract latent capabilities from language models without retraining.

\textbf{Keywords:}
Language Models; Markov Chain Monte Carlo; Simulated Annealing; Power Sampling; Theory of Mind
\end{abstract}

\section{Introduction}

Autoregressive language models have been widely argued to be fundamentally limited \citep{chomsky2023noam, lecun2022path, marcus2022deep, mahowald2024dissociating, wong2023word}. One line of evidence is that such models can produce locally coherent text that nevertheless contradicts earlier commitments (e.g., facts, constraints, or implied states), because no explicit mechanistic constraint enforces behavior consistent with a globally coherent ``world model'' \citep{ha2018world}.

Theory-of-Mind (ToM) tasks provide a particularly stringent test of a computational model’s ability to maintain globally consistent inference \citep{gandhi2023understanding, ullman2023large, strachan2024testing}. ToM requires consistency over a latent belief graph (e.g., Sally believes X; the world is Y; Anne believes that Sally believes Z; etc). If an autoregressive language model fails to tightly bind these latent variables across the full context, a characteristic pattern emerges: the model may answer one question correctly but fail on another that requires the same underlying belief bookkeeping \citep{ullman2023large, muchovej2025large, hu2025re}.

While scaling model size and training data does improve ToM performance, it remains possible to construct novel edge cases on which even frontier models fail to reason correctly \citep{ullman2023large}. We therefore adopt an alternative perspective by focusing instead on small language models (i.e., models with fewer than 4B parameters). We develop test-time optimization methods that sample and evaluate from an autoregressive model’s sequence-level distributions, rather than from the immediate next-token distribution. Sampling at the sequence level enables more coherent evaluation of longer generations, thereby recovering globally consistent signals from the autoregressive model. This, in turn, yields a more holistic assessment of an autoregressive language model’s ToM capabilities, as well as other abilities that require reasoning over extended token sequences governed by a coherent latent graph.

Directly sampling from the sequence-level distribution of an autoregressive model is computationally intractable, because it requires accounting for all possible future token continuations. MCMC samplers address this challenge by maintaining a current sequence and a newly proposed sequence, and comparing their relative likelihoods to evolve a Markov chain, such that the chain converges to a stationary distribution corresponding to the model’s sequence-level distribution \citep{karan2025reasoning}. We leverage this MCMC approach and incorporate a temperature schedule from simulated annealing to convert the sampler into an optimizer over sequences. In principle, adjusting the temperature allows us to control the degree of exploration versus exploitation in the sequence proposals, ultimately enabling mode discovery in the complex sequence-level distribution defined by the autoregressive model.

As a preview, we find that simulated annealing substantially improves the ToM performance of small language models, outperforming other test-time compute methods such as Chain-of-Thoughts \citep[CoT;][]{wei2022chain} and MCMC sampling from the sequence-level distribution \citep{karan2025reasoning}. The sample-then-commit dynamics of mode discovery enabled by simulated annealing unlock substantially stronger capabilities in much smaller models, which were previously thought to be inadequate for many ToM tasks that require maintaining a consistent latent graph. In the following sections, we review related work, describe the proposed method, and present our experiments with small language models.

\section{Background}

\textbf{Limitations of Autoregressive Models.}
A core criticism of autoregressive language models is that autoregressive sampling is inherently greedy: errors can accumulate over time, leading to increasingly incorrect generations because each new token depends only on previously generated tokens \citep{lecun2022path}. In contrast, true world models--used for planning, simulation, and robust ToM--typically include (i) latent state representations that persist independently of surface tokens, (ii) transition dynamics that causally simulate future states, and (iii) policy or utility functions for planning and decision-making \citep{ha2018world, lecun2022path}. As a result, autoregressive language models differ fundamentally from how world models maintain global consistency across states. Their generative mechanism (next-token prediction) differs from the internal world simulation required for planning and reasoning \citep{berglund2023reversal}, making their reasoning fragile.

\textbf{Test-Time Compute Unlocks Advanced Reasoning Abilities.}
An alternative, more optimistic view of autoregressive language models is that they have internalized a substantial amount of commonsense knowledge about the world from pretraining over Internet text \citep{brown2020language}, and that the key challenge lies in test-time elicitation of these latent capabilities \citep{snell2025scaling, shao2024deepseekmath}. From this perspective, both pretraining and test-time computation are equally important for a holistic evaluation of model performance: without appropriate elicitation at test time, researchers may underestimate a model’s true capabilities, even though substantial latent abilities remain to be unlocked.

Note the crucial distinction between test-time compute and post-training methods. The former typically does not involve modifying model parameters; instead, it adaptively reshapes the model’s output distribution at inference time, conditioned on a given prompt. The goal of test-time compute is therefore to elicit more complex distributions from a fixed pretrained language model without retraining. A variety of test-time compute strategies have been explored for autoregressive language models, including CoT \citep{wei2022chain}, Tree-of-Thought \citep{yao2023tree}, self-consistency \citep{qu2024recursive}, and power sampling with MCMC \citep{karan2025reasoning}. Our proposed method falls into this category: it is a test-time optimization approach that improves a model’s reasoning performance without changing the model weights.

\textbf{Theory-of-Mind in Machine Intelligence.}
There is an ongoing debate within the cognitive science community regarding whether current language models possess ToM. While these models can pass ToM tests originally designed for human participants \citep{kosinski2024evaluating}, they often fail to generalize to simple variations of the same test vignettes \citep{ullman2023large}. Researchers on both sides of this debate largely agree that future iterations of language models will continue to pass classical ToM tests and their variants \citep{ullman2023large, hu2025re, kosinski2024evaluating}. However, they disagree on whether success on such benchmark tests constitutes meaningful evidence for attributing genuine ToM capabilities to machine intelligence \citep{block1981psychologism, zhou2023far, amirizaniani2024llms}.

We aim to contribute a fresh perspective to this debate by demonstrating that substantially stronger ToM performance can be recovered from much smaller language models; models that were previously believed to be incapable of solving ToM tasks. By focusing on test-time optimization rather than architectural changes or additional training, our approach helps disentangle competence from elicitation. From a cognitive science perspective, this suggests that apparent ToM failures in language models may sometimes reflect limitations of evaluation and inference procedures rather than an absence of underlying representational capacity. As such, our results encourage a re-examination of what ToM benchmarks measure, and how methodological choices shape conclusions about cognitive capacities in both artificial and natural systems.

\section{Methods}

\begin{figure*}[t!]
    \centering
    \includegraphics[width=\linewidth]{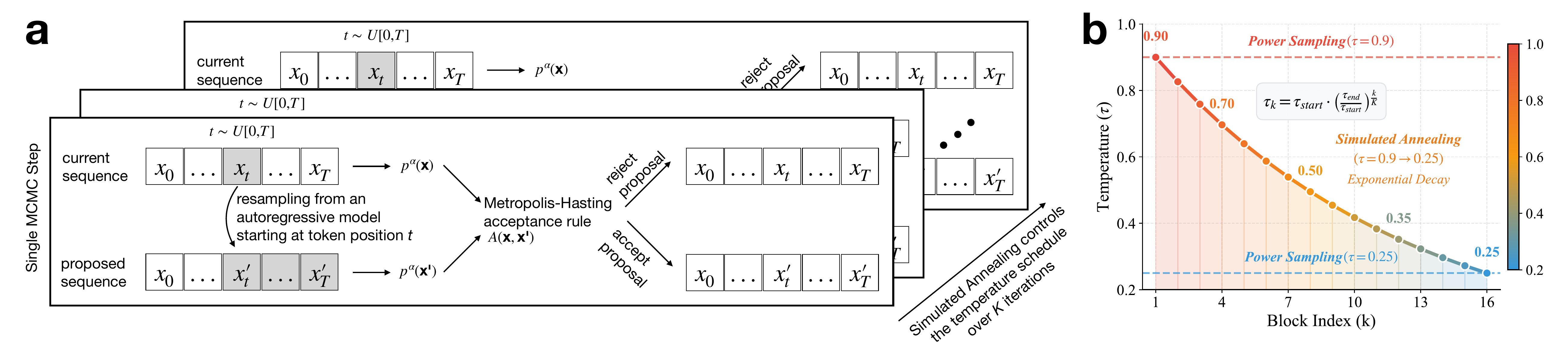}
    \caption{\textbf{Illustration of the simulated annealing approach for optimizing reasoning in autoregressive language models.} \textbf{(a)} Simulated annealing controls the temperature (i.e., the inverse of the power parameter $\alpha$) across successive MCMC iterations. At each iteration $k$, a token position $t$ is selected uniformly at random from the current sequence. The autoregressive model then resamples the sequence suffix conditioned on $\textbf{x}_{<t}$ to form a proposal sequence. The current and proposed sequences are subsequently evaluated using a Metropolis–Hastings acceptance step. \textbf{(b)} The temperature schedule used in our experiments, along with the corresponding fixed-temperature control conditions. }
    \label{fig:SA_illustration}
\end{figure*}

We now introduce our test-time optimization method based on simulated annealing for enhancing reasoning in autoregressive models. First, we review the computational challenges associated with sampling from distorted sequence-level distributions using only autoregressive conditionals. We then discuss MCMC-based techniques that approximately address these challenges, followed by a simulated annealing procedure that converts the MCMC sampler into an optimizer capable of supporting more advanced reasoning.

\textbf{Why Distort Sequence-Level Distributions?}
Autoregressive language models are trained to approximate the true sequence-level distribution \citep{shannon1951prediction, zhang2024should}, but direct sampling from this distribution tends to favor \textit{locally typical} completions rather than \textit{globally coherent} ones \citep{lecun2022path, wong2023word}. Many reasoning tasks, including ToM problems, require selecting sequences that remain globally consistent with an underlying latent structure, even when such sequences are not locally probable at every step \citep{gandhi2023understanding, strachan2024testing, zhu2024incoherent}. Distorting the sequence-level distribution (e.g., by sharpening it) re-allocates probability mass toward globally coherent trajectories. From this perspective, sequence-level distortion functions as a test-time inductive bias that re-prioritizes global consistency at the sequence level and constraint satisfaction over local typicality.

\textbf{Sampling from Distorted Sequence-level Distributions.}
By construction, sampling from the next-token distribution conditional on a prefix is straightforward for autoregressive models. Given a prefix $\textbf{x}_{<t} = (x_0, x_1, \ldots, x_{t-1})$, an autoregressive model directly specifies the distribution over the next token, $p_{\mathrm{LM}}(x_t \mid \textbf{x}_{<t})$. It is also straightforward to sample from the corresponding sequence-level distribution. Consider a sequence of $T+1$ tokens with joint distribution $p(x_0, x_1, \ldots, x_T)$. Using the same autoregressive model, this joint distribution factorizes as a product of next-token distributions,
\begin{align}
    p(x_0, x_1, \ldots, x_T) = \prod_{t=0}^T p_\text{LM}(x_t|\textbf{x}_{<t})
\end{align}
Therefore, sampling tokens sequentially from the autoregressive model directly yields a sample from this joint sequence-level distribution.

What is more challenging is sampling from a distorted (e.g., sharpened) sequence-level distribution while still using the same underlying autoregressive model. One simple way to achieve such sharpening is to sample from a power distribution, $p^\alpha$, in which the joint distribution is exponentiated by a power $\alpha$. When $\alpha > 1$, the distribution is sharpened: the relative weight assigned to higher-likelihood sequences increases, while the relative weight assigned to lower-likelihood sequences decreases. In other words, the original joint distribution becomes more peaked.

Unfortunately, sequentially sampling from the correspondingly distorted autoregressive distribution [i.e., sampling from $p(x_t|\textbf{x}_{<t})^\alpha$] does not, in general, produce samples from the power-sharpened joint distribution. To see why, consider a two-token sequence $(x_0, x_1)$ generated by an autoregressive model:
\begin{align}
    p(x_0, x_1) = p(x_0)p(x_1|x_0)
\end{align}
The power-sharpened joint distribution therefore takes the following form:
\begin{align}
    p(x_0, x_1)^\alpha = p(x_0)^\alpha p(x_1|x_0)^\alpha
\end{align}
To sample $x_0$ from this distribution, we must marginalize over the future token $x_1$:
\begin{align}
    \hat p_\text{correct}(x_0) = \sum_{x_1}p(x_0, x_1)^\alpha = p(x_0)^\alpha \sum_{x_1} p(x_1|x_0)^\alpha \label{eq:correct_conditionals}
\end{align}
Note that the correct marginal $\hat p_\text{correct}(x_0)$ is \textit{not} proportional to $p(x_0)^\alpha$ unless $p(x_1 \mid x_0)$ is identical for all $x_0$, which is almost never the case.

In contrast, naive low-temperature sampling from an autoregressive model samples sequentially from the conditional
\begin{align}
    \hat{p}_\text{naive}(x_0) = \frac{p(x_0)^\alpha}{\sum_{x_0'} p(x_0)^\alpha}
\end{align}
where the temperature $\tau = 1/\alpha$. This conditional ignores the factor $\sum_{x_1} p(x_1 \mid x_0)^\alpha$; that is, it does not account for how many high-probability future continuations a given token admits. In summary, power-sharpening the joint distribution rewards prefixes that lead to many good continuations, whereas power-sharpening the autoregressive conditionals only rewards prefixes that are locally likely.

\textbf{Power Sampling with Autoregressive MCMC.}
As shown in Equation~\ref{eq:correct_conditionals}, direct sampling from the power-sharpened distribution requires normalization over all possible future sequences. Fortunately, MCMC methods directly address this challenge, as they allow sampling without explicit access to the normalizing constant. In particular, the Metropolis–Hastings algorithm provides a convenient framework for constructing such samplers \citep{metropolis1953equation}. In Metropolis–Hastings, a new token sequence $\textbf{x}'$ is proposed from the current sequence $\textbf{x}$ using an arbitrary proposal distribution $q(\textbf{x}' \mid \textbf{x})$. The proposed sequence is then accepted with probability
\begin{align}
    A(\textbf{x}, \textbf{x}') = \min \Big\{ 1, \frac{p^\alpha(\textbf{x})q(\textbf{x}'|\textbf{x})}{p^\alpha(\textbf{x}')q(\textbf{x}|\textbf{x}')}   \Big\} 
\end{align}
If the proposal is rejected, the Markov chain remains at the current state $\textbf{x}$.

Recently, \citet{karan2025reasoning} proposed using the autoregressive distribution of a language model as the proposal distribution $q$ in this MCMC framework (see Figure \ref{fig:SA_illustration}a for illustrations). Under this choice, both $q(\textbf{x}' \mid \textbf{x})$ and $q(\textbf{x} \mid \textbf{x}')$ can be efficiently computed as resampling likelihoods from the autoregressive model. To better integrate with the autoregressive structure, they select a token position uniformly at random (i.e., $t\sim U[0, T]$), use all preceding tokens as a fixed prefix (i.e., $\textbf{x}_{<t}
$), and resample the remaining suffix from the language model. This procedure yields a Power Sampling with Autoregressive MCMC algorithm that can sample from the power-sharpened joint distribution with substantially reduced computational cost \citep{karan2025reasoning}.

\begin{figure*}[t!]
  \begin{center}
    \includegraphics[page=1, width=\linewidth]{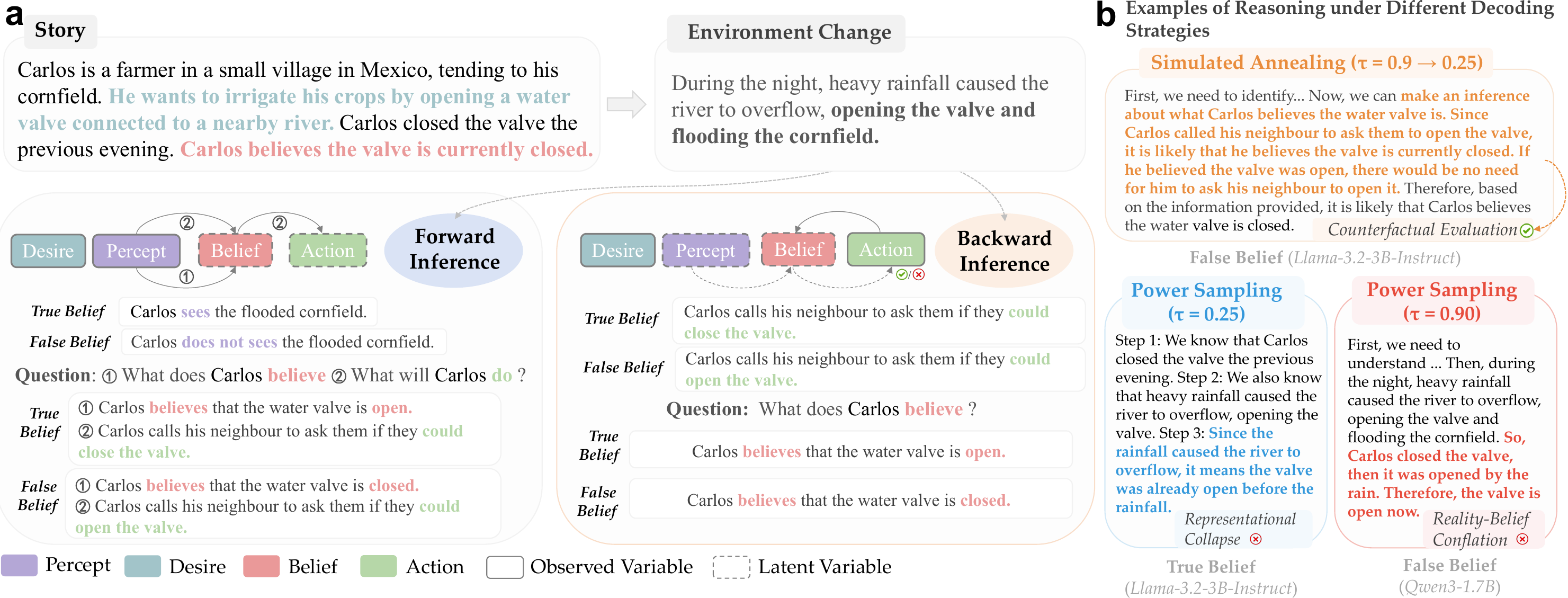}
  \end{center}
    \caption{
    \textbf{(a)} An example ToM narrative from the BigToM benchmark \citep{gandhi2023understanding}. The same narrative supports both forward inference (predicting an agent’s action from its beliefs) and backward inference (inferring an agent’s beliefs from its action). Within each inference type, scenarios are further divided into True Belief (the agent observes the critical event) and False Belief (the agent does not observe the event) conditions.
    \textbf{(b)} Representative examples of model-generated reasoning under different decoding strategies. These examples highlight distinct modes of reasoning corresponding to globally consistent versus locally plausible but inconsistent belief interpretations.
    }
  \label{fig:BigToM_task_and_reasoning_examples}
\end{figure*}

\textbf{Using Simulated Annealing to Optimize Reasoning.}
We now have an MCMC sampler that can draw samples from the power-sharpened joint distribution. However, for certain cognitive tasks, the goal is not to accurately sample from this distribution or to represent its uncertainty, but rather to find a global optimum (e.g., a maximum-a-posteriori solution).

A general strategy for converting an MCMC sampler into an optimization algorithm is simulated annealing \citep{pincus1970monte, kirkpatrick1983optimization}. Annealing is inspired by thermodynamics, in which metals are slowly cooled to reach low-energy crystalline states. Algorithmically, simulated annealing can be viewed as a form of hill-climbing optimization in which, rather than always selecting the best move, the algorithm proposes random moves. Like an MCMC sampler, it preferentially accepts better proposals, but it also accepts worse ones with some probability, allowing it to escape local optima.

We extend the power sampling algorithm proposed by \citet{karan2025reasoning} by incorporating annealing principles. Specifically, we implement a temperature schedule that starts at a high value and gradually decreases to a low value, analogous to simulated annealing (see Figure \ref{fig:SA_illustration}b). An initially high temperature allows the MCMC sampler to more effectively explore the joint distribution, whereas the subsequent cooling phase encourages convergence toward a mode solution.

It is important to note that under simulated annealing, the chain is no longer stationary, because detailed balance is violated over the course of the temperature schedule \citep{kirkpatrick1983optimization}. In other words, unlike the power sampling algorithm \citep{karan2025reasoning}, samples generated during annealing are not drawn from any fixed target distribution. This is appropriate for optimization, where mode discovery is the priority, but not for treating the samples as posterior draws.

\textbf{Implementational Details.}
In power sampling, the sharpness of the effective sampling distribution is controlled by the power exponent $\alpha$, which is inversely related to the temperature $\tau$ via $\tau = 1 / \alpha$. 
We implement simulated annealing using an exponentially decaying temperature schedule, starting at a high temperature ($\tau = 0.90$) and gradually decreasing to a low temperature ($\tau = 0.25$) over Metropolis–Hastings iterations. Specifically, the temperature at iteration $k$ is defined as
\begin{align}
\tau_k = \tau_{\text{start}} \left( \frac{\tau_{\text{end}}}{\tau_{\text{start}}} \right)^{k / K}
\end{align}
where $\tau_{\text{start}} = 0.90$, $\tau_{\text{end}} = 0.25$, $k$ indexes the current iteration, and $K$ denotes the total number of sampling steps (see Figure \ref{fig:SA_illustration}b for illustration). This exponential schedule promotes rapid exploration early in sampling, followed by progressively stronger concentration of probability mass in the sequence-level distribution.


Following the power sampling algorithm \citep{karan2025reasoning}, we exploit the structure of autoregressive generation to accelerate MCMC by generating sequences block by block rather than token by token, with each iteration resampling a block of tokens. All other hyperparameters are fixed to match prior work \citep{karan2025reasoning}: maximum new tokens = 512, number of blocks = 16, and number of MCMC steps = 10.


\begin{figure*}[t!]
  \begin{center}
    \includegraphics[page=1, width= \linewidth]{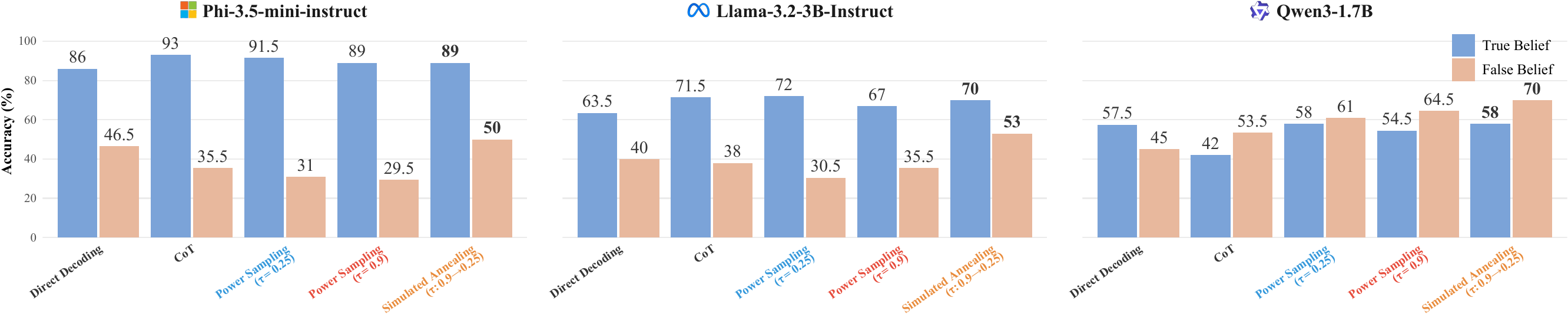}
  \end{center}
    \caption{\textbf{Accuracy across small language models and decoding strategies.} Panels (left to right) correspond to \texttt{Phi-3.5-Mini-Instruct}, \texttt{LLaMA-3.2-3B-Instruct}, and \texttt{Qwen3-1.7B}. Within each panel, bars (left to right) represent five decoding strategies: direct decoding, Chain-of-Thought (CoT), low-temperature power sampling, high-temperature power sampling, and simulated annealing. Results are shown separately for True Belief (TB) and False Belief (FB) conditions in backward inference. \textcolor{blue}{Blue} bars indicate TB accuracy, while \textcolor{orange}{orange} bars indicate FB accuracy. }
  \label{fig:ToM_model_accuracy}
\end{figure*}

\section{Experiments}

\textbf{Small Language Models.}
We evaluate three open-source small language models: \texttt{Phi-3.5-Mini-Instruct} (3.8B parameters), \texttt{LLaMA-3.2-3B-Instruct}, and \texttt{Qwen3-1.7B}. These models were chosen to span different model families and development groups while remaining within the regime of small autoregressive models.

\textbf{Evaluating ToM Performance.}
All models are evaluated using the BigToM benchmark \citep{gandhi2023understanding}, which extends classical ToM paradigms from cognitive science into a scalable, text-based evaluation framework suitable for modern language models. The benchmark places particular emphasis on belief tracking, belief updating, and generalization under controlled variations. BigToM consists of synthetic narrative vignettes involving multiple agents, objects, and events, each followed by structured questions probing mental states. Rather than scoring isolated responses, BigToM assesses whether a model maintains a coherent latent belief graph across multiple queries, thereby directly targeting a core weakness of autoregressive models.

We evaluate all 200 narrative templates in the BigToM benchmark \citep{gandhi2023understanding}, focusing on backward-inference cases involving true and false beliefs, as small language models already achieve near-ceiling performance on other ToM inference tasks. This yields a total of 400 test instances (see Figure \ref{fig:BigToM_task_and_reasoning_examples}a for an example). Each story comprises the following elements: (1) an agent with a goal or desire, (2) an initial belief state, (3) a critical environmental change that is either observed (True Belief; TB) or unobserved (False Belief; FB) by the agent, (4) a subsequent action taken by the agent, and (5) a belief query requiring a binary choice between maintaining the initial belief or updating it in response to the environmental change.

For each model, we compare five decoding strategies: (i) direct decoding, in which the model produces an answer directly from its autoregressive conditionals; (ii) CoT decoding using a standard step-by-step reasoning prompt; (iii) low-temperature power sampling ($\tau = 0.25$); (iv) high-temperature power sampling ($\tau = 0.90$); and (v) Simulated annealing with a temperature schedule $\tau = 0.90 \rightarrow 0.25$. For the direct and CoT baselines, decoding is greedy, with the temperature set to zero (see Figure \ref{fig:BigToM_task_and_reasoning_examples}b for examples).

\textbf{Results.} 
Because all three small language models achieve ceiling performance on forward-inference tasks under direct decoding, we focus our analysis on backward-inference tasks. Figure~\ref{fig:ToM_model_accuracy} summarizes ToM performance across five decoding conditions, three autoregressive language models, and two belief types (i.e., TB and FB). Results are reported separately for TB and FB conditions. 

When autoregressive language models use direct decoding to solve backward-inference tasks, we observe an asymmetry in accuracy between TB and FB conditions. Models perform better on TB tasks (with accuracy ranging from 57.5\% to 86\%), whereas performance deteriorates markedly on FB tasks (with accuracy ranging from 40\% to 46.5\%). In fact, the highest FB accuracy is lower than the lowest TB accuracy in our evaluations. This gap replicates prior findings that FB reasoning poses a greater challenge for language models \citep{gandhi2023understanding, strachan2024testing}. This pattern also parallels findings in the cognitive development literature showing that children reliably perform worse on FB tasks than on structurally analogous TB controls, reflecting the greater cognitive demands of representing another’s belief when it diverges from reality \citep{gopnik1988children, baron1985does}. That is, FB reasoning appears to be a computationally harder problem both for small language models and for human learners.

For both the \texttt{Phi-3.5} and \texttt{LLaMA-3.2} models, CoT and power sampling with fixed temperatures ($\tau = 0.25$ or $\tau = 0.90$) tend to improve accuracy on TB tasks while simultaneously degrading performance on FB tasks. In contrast, for the \texttt{Qwen3} model, the same decoding strategies often improve FB accuracy without affecting TB performance.

Importantly, higher accuracy under these decoding strategies does not automatically correspond to improved ToM reasoning quality. Qualitative inspection of model-generated completions reveals that apparent gains in accuracy can be driven by errors rather than coherent ToM inference. For example, in the TB condition using low-temperature power sampling for \texttt{LLaMA-3.2-3B-Instruct} (Figure~\ref{fig:BigToM_task_and_reasoning_examples}b), the model exhibits a severe form of representational collapse: it correctly states that ``rainfall caused the river to overflow, opening the valve,'' but subsequently concludes that the valve must have been open prior to the rainfall. This inconsistency arises at the level of world-state representation, before any belief attribution is required, indicating a failure to maintain a coherent causal model of the world.

Similarly, even in cases where \texttt{Qwen3-1.7B} attains relatively high FB accuracy (up to 64.5\% under high-temperature power sampling), the underlying reasoning can conflate objective reality with the agent’s belief state (see Figure~\ref{fig:BigToM_task_and_reasoning_examples}b). In these model-generated completions, changes to the true state of the world are implicitly treated as automatic updates to the agent’s beliefs, disregarding the agent’s informational limitations.

When considering both TB and FB conditions jointly, simulated annealing outperforms all other decoding strategies. Notably, these gains are most pronounced in the more challenging FB condition, in which the agent does not observe the critical event. Beyond accuracy improvements, simulated annealing also yields qualitatively stronger reasoning. Model-generated completions exhibit belief-sensitive inference patterns characteristic of correct ToM reasoning, including explicit references to perceptual access (e.g., whether the agent is aware or unaware of critical events), a clear separation between perception and belief, and systematic inference from observed actions back to the agent’s latent belief state.

Notably, simulated annealing often elicits reasoning chains resembling counterfactual evaluation. In these cases, the model explicitly considers alternative belief hypotheses and assesses their plausibility by examining the actions that would be expected under each belief. For example, in the FB case illustrated in Figure~\ref{fig:BigToM_task_and_reasoning_examples}b, the model reasons that if Carlos believed the valve were already open, calling a neighbor to ask them to open it would be unnecessary; because he does make that request, the model infers that Carlos likely believes the valve is closed. This form of belief hypothesis testing closely mirrors inverse planning strategies commonly associated with ToM reasoning. Such structured reasoning is largely absent under other decoding strategies, where completions tend to conflate belief with objective reality or lack a coherent latent representation.

\section{Discussion}

We introduced a simulated annealing method to elicit stronger ToM reasoning abilities from autoregressive language models without additional training or external verifiers. Substantial improvements in ToM performance can indeed be achieved by imposing a temperature schedule on an MCMC sampler that draws samples from temperature-distorted sequence-level distributions of the autoregressive model. The fact that no additional training or verification is required suggests that the representational capacity of pretrained autoregressive language models is considerably more powerful than is typically assumed.

\textbf{Towards A Unifying View of Test-Time Compute.}
The stronger performance achieved with simulated annealing relative to direct decoding and CoT suggests that principled, sampling-based approaches are a promising direction for understanding decoding and test-time compute. Indeed, many test-time compute strategies have already been interpreted through the lens of modifying a model's sequence-level distribution at test time \citep{snell2025scaling}. Our results indicate that the sequence-level distribution defined by an autoregressive model is an appropriate target for better reasoning. Viewing test-time inference as mode discovery over this complex sequence-level distribution may help unify existing strategies and inspire new ones. Simulated annealing represents one such strategy, enabling the model to more effectively traverse the sequence-level probabilistic landscape and identify higher-quality modes.

\textbf{Connections to Sampling-based Models of Human Cognition.}
Much of cognitive theory assumes that human behavior approximates Bayesian inference via sampling-based processes, such as MCMC, and that systematic deviations from Bayesian-optimal solutions can be explained by assuming a limited number of posterior samples \citep{griffiths2012bridging, gershman2009perceptual, zhu2024autocorrelated, bramley2017formalizing}. Sampling from an LLM’s sequence-level distribution offers a novel avenue for modeling human cognition by directly connecting to these established sampling-based accounts. This perspective is also intuitively appealing: during the short timescale of laboratory experiments, human participants are unlikely to undergo substantial parameter changes, closely resembling test-time inference in pretrained language models, where model weights remain fixed but behavior can adapt across different levels of performance.

\textbf{Limitations and Future Research.}
Despite the substantial gains achieved by simulated annealing at test time, several limitations remain. Even under annealing, failures persist in cases requiring deeply nested belief reasoning or complex causal structures, indicating that sequence-level optimization can reveal, but not fully overcome, limitations in the fixed representations of pretrained language models.

Our method also incurs additional inference-time computational cost. Although simulated annealing is more efficient than exhaustive search, it is significantly more expensive than standard autoregressive decoding or single-pass CoT prompting, raising scalability concerns for longer sequences, larger models, or real-time applications. Future work should investigate more efficient proposal mechanisms, adaptive stopping criteria, or hybrid approaches that selectively apply sequence-level optimization.

Finally, it remains an open question whether the benefits of sequence-level distortion generalize to other reasoning domains, such as moral reasoning, strategic interaction, multi-agent coordination, or long-horizon planning. While simulated annealing is an effective strategy for navigating complex sequence-level distributions, it is unlikely to be unique. Framing test-time inference as mode discovery over a high-dimensional sequence space suggests a broader family of optimization-based decoding methods.

\printbibliography

\end{document}